\documentclass[letterpaper]{article} 
\usepackage{aaai23}  
\usepackage{times}  
\usepackage{helvet}  
\usepackage{courier}  
\usepackage[hyphens]{url}  
\usepackage{graphicx} 
\usepackage{amssymb}
\usepackage{epsfig}
\usepackage{subfig}
\usepackage{amsmath}
\usepackage{mathrsfs} 
\usepackage{natbib}  
\usepackage{caption} 
\usepackage{algorithm}
\usepackage{algpseudocode}
\usepackage{newfloat}
\usepackage{listings}
\DeclareCaptionStyle{ruled}{labelfont=normalfont,labelsep=colon,strut=off} 
\floatstyle{ruled}
\newfloat{listing}{tb}{lst}{}
\floatname{listing}{Listing}
\title{Improving Adversarial Robustness with Self-Paced Hard-Class Pair Reweighting}
\author {
    Pengyue Hou,\textsuperscript{\rm 1}
    Jie Han, \textsuperscript{\rm 1}
    Xingyu Li \textsuperscript{\rm 1}
}
\affiliations {
    \textsuperscript{\rm 1} University of Alberta\\
    pengyue@ualberta.ca, jhan8@ualberta.ca, xingyu@ualberta.ca
}

\usepackage{bibentry}

\begin{document}

\maketitle

\begin{abstract} 
Deep Neural Networks are vulnerable to adversarial attacks. Among many defense strategies, adversarial training with untargeted attacks is one of the most effective methods. Theoretically, adversarial perturbation in untargeted attacks can be added along arbitrary directions and the predicted labels of untargeted attacks should be unpredictable. However, we find that the naturally imbalanced inter-class semantic similarity makes those hard-class pairs become virtual targets of each other. This study investigates the impact of such closely-coupled classes on adversarial attacks and develops a self-paced reweighting strategy in adversarial training accordingly. Specifically, we propose to upweight hard-class pair losses in model optimization, which prompts learning discriminative features from hard classes. We further incorporate a term to quantify hard-class pair consistency in adversarial training, which greatly boost model robustness. Extensive experiments show that the proposed adversarial training method achieves superior robustness performance over state-of-the-art defenses against a wide range of adversarial attacks. The code of the proposed SPAT is published at \textit{https://github.com/puerrrr/Self-Paced-Adversarial-Training}.
\end{abstract}

\section{Introduction}

In recent years, DNNs are found to be vulnerable to adversarial attacks, and extensive work has been carried out on how to defend or reject the threat of adversarial samples~\cite{szegedy2013intriguing, goodfellow2014explaining, nguyen2015deep}. Adversarial samples are carefully generated with human-imperceptible noises, yet they can lead to large performance degradation of well-trained models. 

While numerous defenses have been proposed, adversarial training (AT) is a widely recognized strategy~\cite{madry2017towards} and achieves promising performance against a variety of attacks. AT treats adversarial attacks as an augmentation method and aims to train models that can correctly classify both adversarial and clean data. Based on the AT framework, further robustness improvements can be achieved by exploiting unlabeled, miss-classified data, pre-training, etc~\cite{alayrac2019labels, carmon2019unlabeled, hendrycks2019using, zhai2019adversarially, wang2019improving, jiang2020robust, fan2021does, Hou2022}.

\begin{figure}[t]
  \centering
    \subfloat[
    PDG attacks on vanilla-trained model] {\includegraphics[width=0.23\textwidth]{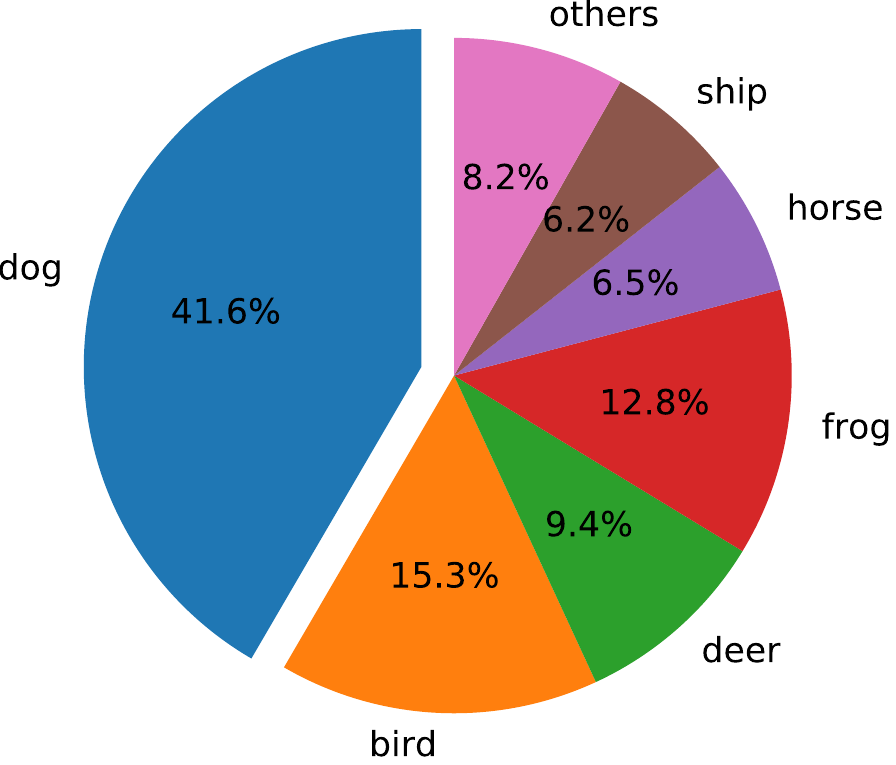}}
    \hspace{0.1cm}
  \subfloat[PDG attacks on SPAT-trained model]{\includegraphics[width=0.23\textwidth]{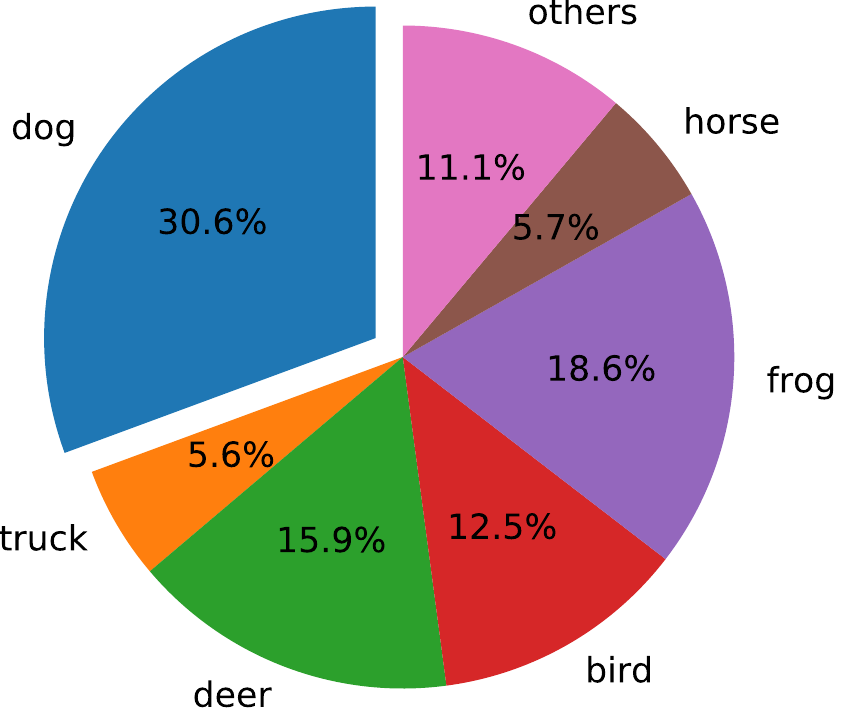}}
  \caption{Predictions of untargeted adversarial attacks (PGD-20) by CIFAR-10 vanilla-trained and SPAT-trained classifiers. (a) For the vanilla-trained model, over 40\% of the dog images are misclassified as cats and (b) it is reduced to 30.6\% with the SPAT-trained model.} 
  \label{cat dog bias}
\end{figure}

In existing adversarial training, untargeted attacks are widely used in model optimization and evaluation~\cite{moosavi2016deepfool, madry2017towards, zhang2019theoretically, wang2019improving, kannan2018adversarial, shafahi2019adversarial, wong2020fast}. Unlike targeted attacks that aim to misguide a model to a particular class other than the true one, untargeted adversaries do not specify the targeted category and perturb the clean data so that its prediction is away from its true label. In theory, adversarial perturbation in untargeted attacks can be added along arbitrary directions and classification of untargeted attacks should be unpredictable. However, the study by Carlini \textit{et al} argues that an untargeted attack is simply a more efficient method of running a targeted attack for each target and taking the \textit{closest} \cite{carlini2017towards}. Figure~\ref{cat dog bias} (a) presents the misclassification statistics of PDG-attacked dog images, where almost half of dog images are misclassified as cats, and over 40\% of the cat images are misclassified as dogs. Considering that cat and dog images share many common features in vision, 
we raise the following questions: \\
\textit{"Does the unbalanced inter-class semantic similarity lead to the non-uniformly distributed misclassification statistics? If \textbf{yes}, are classification predictions of untargeted adversaries predictable?" }

To answer these questions, this paper revisits the recipe for generating gradient-based first-order adversaries and surprisingly discovers that untargeted attacks may be targeted. In theory, we prove that adversarial perturbation directions in untargeted attacks are actually biased toward the hard-class pairs of the clean data under attack. Intuitively, semantically-similar classes constitute \textbf{hard-class pairs (HCPs)} and semantically-different classes form \textbf{easy-class pairs (ECPs)}. 

Accordingly, we propose explicitly taking the inter-class semantic similarity into account in AT algorithm design and develop a self-paced adversarial training (SPAT) strategy to upweight hard/easy-class pair losses and downweight easy-class pair losses, encouraging the training procedure to neglect redundant information from easy class pairs. Since HCPs and ECPs may change during model training (depending on the current optimization status), their scaling factors are adaptively updated at their own pace. Such self-paced reweighting offers SPAT more optimization flexibility. In addition, we further incorporate an HCP-ECP consistency term in SPAT and show its effectiveness in boosting model adversarial robustness. Our main contributions are:
\begin{itemize}
    \item We investigate the cause of the unevenly distributed misclassification statistics in untargeted attacks. We find that adversarial perturbations are actually biased by targeted sample's hard-class pairs.
    \item We introduce a SPAT strategy that takes inter-class semantic similarity into account. Adaptively upweighting hard-class pair loss encourages discriminative feature learning. 
    \item We propose incorporating an HCP-ECP consistency regularization term in adversarial training, which boosts model adversarial robustness by a large margin.
\end{itemize}

\section{Related Work}
\subsection{Adversarial Attack and Defense}
The objective of adversarial attacks is to search for human-imperceptible perturbation $\boldsymbol{\delta}$ so that the adversarial sample 
\begin{equation}
  \boldsymbol{x'} = \boldsymbol{x} + \boldsymbol{\delta}
\end{equation}
can fool a model $f(\boldsymbol{x};\boldsymbol{\phi})$ well-trained on clean data $\boldsymbol{x}$. Here $\phi$ represents the trainable parameters in a model. For notation simplification, we use $f(\boldsymbol{x})$ to denote $f(\boldsymbol{x};\boldsymbol{\phi})$ in the rest of the paper.  One main branch of adversarial noise generation is the gradient-based method, such as the Fast Gradient Sign Method (FGSM)~\cite{goodfellow2014explaining}, and its variants ~\cite{kurakin2016adversarial, madry2017towards}. Another popular strategy is optimization based, such as the CW attack~\cite{carlini2017towards}.

Several pre/post-processing-based methods have shown outstanding performance in adversarial detection and classification tasks ~\cite{grosse2017statistical, metzen2017detecting,xie2017mitigating, feinman2017detecting, li2017adversarial}. They aim to use either a secondary neural network or random augmentation methods, such as cropping, compression and blurring to strengthen model robustness. However, Carlini \textit{et al.} showed that they all can be defeated by a tailored attack~\cite{carlini2017adversarial}. Adversarial Training, on the other hand, uses regulation methods to directly enhance the robustness of classifiers. Such optimization scheme is often referred to as the "min-max game":
\begin{equation}
  \underset{\boldsymbol{\phi}}{\mathrm{argmin }}
  \mathbb{E}_{(\boldsymbol{x},\boldsymbol{y})\sim D}[\underset{\delta \in S}{\mathrm{max }} \mathscr{L}(f(\boldsymbol{x'}),\boldsymbol{y})],
  \label{minmax}
\end{equation}
where the inner max function aims to generate efficient and strong adversarial perturbation based on a specific loss function $\mathscr{L}$, and the outer min function optimizes the network parameters $\boldsymbol{\phi}$ for model robustness. Another branch of AT aims to achieve \textit{logit level robustness}, where the objective function not only requires correct classification of the adversarial samples, but also encourages the logits of clean and adversarial sample pairs to be similar ~\cite{kannan2018adversarial, zhang2019theoretically, wang2019improving}. Their AT objective functions usually can be formulated as a compound loss:
\begin{equation}
\mathscr{L}(\boldsymbol{\theta}) = \mathscr{L}_{acc}+\lambda\mathscr{L}_{rob}
\end{equation}
where $\mathscr{L}_{acc}$ is usually the cross entropy (CE) loss on clean or adversarial data, $\mathscr{L}_{rob}$ quantifies clean-adversarial logit pairing, and $\lambda$ is a hyper-parameter to control the relative weights for these two terms. The proposed SPAT in this paper introduces self-paced reweighting mechanisms upon the compound loss and soft-differentiates hard/easy-class pair loss in model optimization for model robustness boost.

\subsection{Re-weighting in Adversarial Training}
Re-weighting is a simple yet effective strategy for addressing biases in machine learning, for instance, class imbalance. When class imbalance exists in the datasets, the training procedure is very likely over-fit to those categories with a larger amount of samples, leading to unsatisfactory performance regarding minority groups. With the re-weighting technique, one can down-weight the loss from majority classes and obtain a balanced learning solution for minority groups. 

Re-weighting is also a common technique for hard example mining. Generally, hard examples are those data that have similar representations but belong to different classes. Hard sample mining is a crucial component in deep metric learning~\cite{hoffer2015deep, hermans2017defense} and Contrastive learning~\cite{chen2020simple,khosla2020supervised}. With re-weighting, we can directly utilize the loss information during training and characterize those samples that contribute large losses as hard examples. For example, OHEM~\cite{shrivastava2016training} and Focal Loss~\cite{lin2017focal} put more weight on the loss of misclassified samples to effectively minimize the impact of easy examples. 

Previous studies show that utilizing hard adversarial samples promotes stronger adversarial robustness~\cite{madry2017towards, wang2019improving, mao2019metric, pang2020boosting}. For instance, MART~\cite{wang2019improving} explicitly applies a re-weighting factor for misclassified samples by a soft decision scheme. Recently, several re-weighting-based algorithms have also been proposed to address fairness-related issues in AT. \cite{wang2021imbalanced} adopt a re-weighting strategy to address the data imbalance problem in AT and showed that adversarially trained models can suffer much worse performance degradation in under-represented classes.
Xu \textit{et al.}~\cite{xu2021robust} empirically showed that even in balanced datasets, AT still suffers from the fairness problem, where some classes have much higher performance than others. They propose to combine re-weighting and re-margin for different classes to achieve robust fairness. Zhang \textit{et al.}~\cite{zhang2020geometry} propose to assign weights based on how difficult to change the prediction of a natural data point to a different class. 
However, existing AT re-weighting strategies only considered intra-class or inter-sample relationships, but ignored the inter-class biases in model optimization. We propose to explicitly take the inter-class semantic similarity into account in the proposed SPAT strategy and up-weights the loss from hard-class pairs in AT.

\section{Untargeted Adversaries are Targeted}

\begin{figure}[t]
\centerline{\includegraphics[width=0.8\columnwidth]{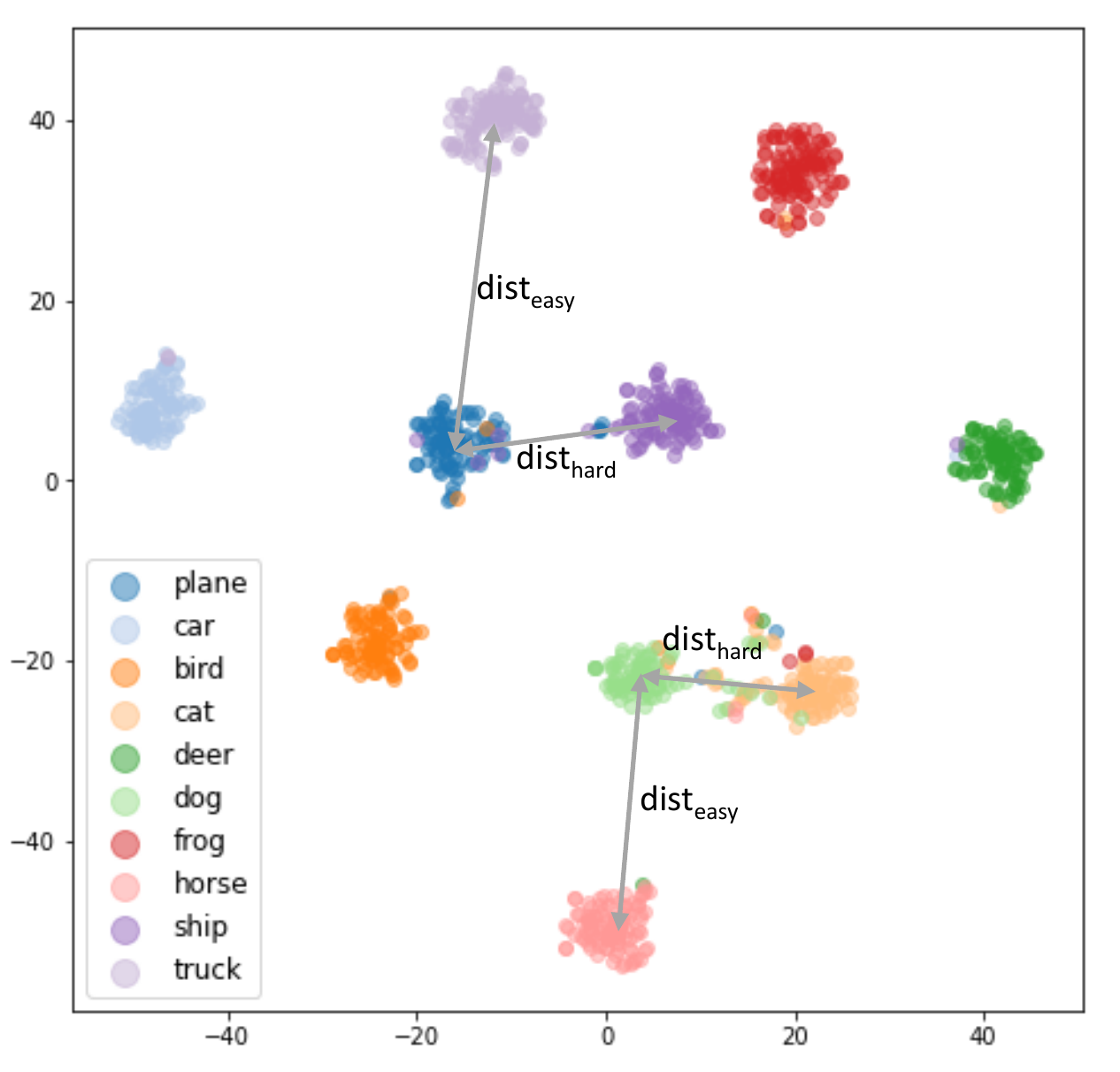}}
\caption{t-SNE visualization of 1000 randomly sampled image embeddings from CIFAR-10. Due to the naturally imbalanced semantic similarity, inter-class distance is much smaller for hard-class pairs.}
\label{stne}
\end{figure}

Untargeted adversarial attacks are usually adopt in adversarial training. In theory, adversarial perturbation in untargeted attacks can be added along arbitrary directions, leading to unpredictable false classification. However, our observations on many adversarial attacks contradict this. For example, when untargeted adversaries attack images of cats, the resulting images are likely to be classified as dogs empirically.
We visualize image embeddings from the penultimate layer of the vanilla-trained model via t-SNE in Figure~\ref{stne}. In the figure, the embeddings of dog and cat images are close to each other, which suggests the semantic similarity in their representations. With this observation, we hypothesize that the unbalanced inter-class semantic similarity leads to the non-uniformly distributed misclassification statistics.

In this section, we investigate this interesting yet overlooked aspect of adversarial attacks and find that untargeted adversarial examples may be highly biased by their hard-class pairs. The insight in this section directly motivates the proposed self-paced adversarial training for model robustness improvement.

\subsection{Notations}
Given a dataset with labeled pairs $\{\mathscr{X,Y}\}=\{(x,y)|x\in \mathbb{R}^{c\times m \times n}, y \in [1,C]\}$, a classifier can be formulated as a mapping function $f: \mathscr{X}\xrightarrow{}\mathscr{Y}$:
\begin{equation}
    f(x)=\mathbb{S}(\boldsymbol{W}^T \boldsymbol{z_x}),
    \label{eqn:classification}
\end{equation}
where $C$ is the number of categories, and $\mathbb{S}$ represents the softmax function in the classification layer. We use $\boldsymbol{z_x}$ to denote the representation of an input sample $x$ in the penultimate layer of the model and $\boldsymbol{W}=(\boldsymbol{w_i},\boldsymbol{w_2},...,\boldsymbol{w_C})$ for the trainable parameters (including weights and bias) of the softmax layer. Note that $\boldsymbol{w_i}$ can be considered as the prototype of class $i$ and the production $\boldsymbol{W}^T\boldsymbol{z_x}$ in (\ref{eqn:classification}) calculates the similarity between $\boldsymbol{z_x}$ and different class-prototype $\boldsymbol{w_i}$. During training, the model $f$ is optimized to minimize a specific loss $\mathscr{L}(f(x),y)$.

In literature, the most commonly used adversarial attacks, such as PGD and its variants, generate adversaries based on first-order derivative information about the network~\cite{madry2017towards}. Such adversarial perturbations can be generally formulated as follows:
\begin{equation}
    x' = x +\epsilon g(\nabla_{\boldsymbol{x}}\mathscr{L}(f(x),y)),
\end{equation}
where $\epsilon$ is the step size to modify the data and $\nabla_{\boldsymbol{x}}$ is the gradient with respect to the input $x$. We take $g$ to denote any function on the gradient, for example, $g(x)=\lVert x \rVert_p$ is the $\ell_p$ norm.


\subsection{Bias in Untargeted Adversarial Attacks}\label{biased AT}
The first-order adversarial attacks usually deploy the CE loss between the prediction $f(x)$ and the target $y$ to calculate adversarial perturbations. The CE loss can be formulated as
\begin{equation}
\mathscr{L}(f(x),y)=-log\frac{e^{\boldsymbol{w_{i}}^{T}\boldsymbol{z_x}}}{\sum_{j=1}^{C}e^{\boldsymbol{w_{j}}^{T}\boldsymbol{z_x}}}
\end{equation}
For notation simplification in the rest of this paper, we have $\sigma (\boldsymbol{w_{i}}^{T}\boldsymbol{z_x}) = \frac{e^{\boldsymbol{w_{i}}^{T}\boldsymbol{z_x}}}{\sum_{j=1}^{C}e^{\boldsymbol{w_{j}}^{T}\boldsymbol{z_x}}}$.

\subsubsection{Lemma 1}(proof in Appendix): \textit{For an oracle model that predicts the labels perfectly on clean data, the gradient of the CE loss with respect to sample $x$ from the $i^{th}$ category is:}
\begin{equation}
\nabla_{\boldsymbol{x}}\mathscr{L}(f(x),y)=[\sum_{j\neq i}^{C}\sigma(\boldsymbol{w_{j}}^{T}\boldsymbol{z_x})\boldsymbol{w_{j}}]\nabla_{\boldsymbol{x}}\boldsymbol{z_x}.
\label{eqn:Lemma1}
\end{equation}

Lemma 1 indicates that for a clean data $x$ from the $i^{th}$ category, its first-order adversarial update follows the direction of the superposition of all false-class prototypes $\boldsymbol{w_{j}}$ for $j\in[1,C], j\neq i$. The weight of the $j^{th}$ prototype $\boldsymbol{w_{j}}$ in the superposition is $\sigma(\boldsymbol{w_{j}}^{T}\boldsymbol{z_x})$. The greater the value of the dot product $\sigma(\boldsymbol{w_{j}}^{T}\boldsymbol{z_x})$, the more bias in adversarial perturbations toward the $i^{th}$ category. In an extreme case where only one $\sigma(\boldsymbol{w_{k}}^{T}\boldsymbol{z_x})$ is non-zero, the untargeted attack becomes a targeted attack.
 
To investigate if the values of $\sigma(\boldsymbol{w_{j}}^{T}\boldsymbol{z_x})$ is equal or not, we let $v_j=\lVert \boldsymbol{w_{j}} \rVert_2$ and $s=\lVert \boldsymbol{z_{x}} \rVert_2$ be the Euclidean norm of the weight and data embedding. Then (\ref{eqn:Lemma1}) in Lemma 1 can be rewritten as $\nabla_{\boldsymbol{x}}\mathscr{L}(f(x),y)=[\sum_{j\neq i}^{C}\sigma(v_js \cos(\boldsymbol{\theta}_{j}))\boldsymbol{w_{j}}]\nabla_{\boldsymbol{x}}\boldsymbol{z_x}$,
where $\cos(\boldsymbol{\theta}_{j})$ measures the angle between the two vectors $\boldsymbol{w_{j}}$ and $\boldsymbol{x_{z}}$. Here, we discussed two conditions.

\subsubsection{Condition 1}. We regulate $v_j=1$ and thus convert the CE loss to the normalized cross entropy (NCE) loss in Lemma 1. Recently, many studies show that NCE loss encourages a model to learn more discriminative features~\cite{wang2018cosface, liu2017sphereface, schroff2015facenet}. Furthermore, such hypersphere embedding boosts adversarial robustness \cite{pang2020boosting}. When we follow NCE's regularization and enforce $v_j=1$, (\ref{eqn:Lemma1}) in Lemma 1 is further simplified to
\begin{equation}
\nabla_{\boldsymbol{x}}\mathscr{L}(f(x),y)=[\sum_{j\neq i}^{C}\sigma(s \cos(\boldsymbol{\theta}_{j}))\boldsymbol{w_{j}}]\nabla_{\boldsymbol{x}}\boldsymbol{z_x},
\label{eqn:Lemma1_2}
\end{equation}
Since $\sigma()$ is a monotonically increasing function, the adversarial update direction is significantly biased by large $\cos(\boldsymbol{\theta}_{j})$. It is noteworthy that $s\cos(\boldsymbol{\theta}_{j})$ quantifies the projection of a data representation $\boldsymbol{x_{z}}$ onto the $j^{th}$ class prototype $\boldsymbol{w_{j}}$, which reflects the inter-class similarity between $z_x$ and a specific false-class prototype. Therefore, this paper defines the false classes associated with a higher $\cos(\boldsymbol{\theta}_{j})$ as the \textbf{hard-class pairs} of data $x$; contrastively, the false classes with large $\boldsymbol{\theta}_{j}$ as the \textbf{easy-class pairs}. With this context, we conclude that the adversarial perturbations introduced by the NCE loss are dominated by those \textbf{hard} classes with smaller inter-class distances from the true data category.

\subsubsection{Condition 2.} We relax the condition $v_j=1$ and extend our discovery to a generic CE loss. Though $v_j$ can be any value in theory, we empirically find that their values are quite stable and even for all $j$ (as shown in Appendix). With these observations, we conclude that untargeted adversaries are actually targeted; Furthermore, the virtual targeted categories are its hard-class pairs.

\begin{figure}[t]
\centerline{\includegraphics[width=0.8
\columnwidth]{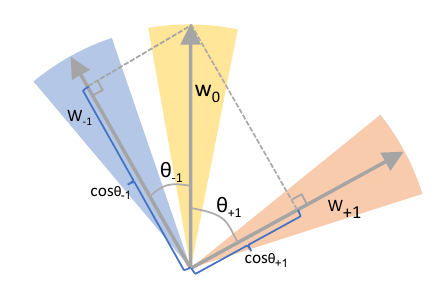}}
\caption{A geometric interpretation of our discovery about untargeted attacks. Different colors represent different classes and $W_i$ is the prototype vector for class i. According to Lemma 1, the overall attack direction for class 0 will be dominated by class -1.}
\label{geometry}
\end{figure}

Figure~\ref{geometry} illustrates a geometric interpretation of our discovery in a simple triplet classification setting, with $y = \{-1, 0, 1\}$. We assume the latent representation of class -1 is closer to class 0 (a hard class pair) and class 1 is farther from class 0 (an easy class pair). Since $cos(\boldsymbol{\theta}_{-1}) > cos(\boldsymbol{\theta}_{+1})$, The attack direction of samples from class 0 is dominated by class -1. Therefore, the data from class 0 is adversarially modified towards class -1.

\section{Self-Paced Adversarial Training} \label{SPAT}
Our discovery in Section \ref{biased AT} motivates the innovation of our re-weighting strategy in the proposed SPAT in twofold.
\begin{itemize}
    \item From the perspective of learning robust, discriminative features. Compared to adversaries from hard-class pairs having similar semantic representations, easy-class pairs contribute less to model optimization. Encouraging a model to learn HCP samples facilitates the model to extract good features.
    \item From the perspective of adversarial defense of untargeted attacks. Thanks to the discovered targeted property of untargeted attacks, we know that many clean data are adversarially modified toward their hard-class pairs. With this prior knowledge of untargeted attacks, one can improve models' robustness by learning HCP adversaries in AT.
\end{itemize}
With the above considerations, our self-pace strategy proposes to up-weights training sample's hard-class pair loss in adversarial training. 

Specifically, following prior arts in adversarial training, the proposed SPAT strategy adopts a compound loss:
\begin{equation}
    \mathscr{L}^{SPAT} = \mathscr{L}_{acc}^{sp} + \lambda \mathscr{L}_{rob}^{sp}
\end{equation}
where $\lambda$ is the trade-off parameter for the accuracy and robustness terms. Notably, we introduces distinct up-weighting policies in $\mathscr{L}_{acc}^{sp}$ and $\lambda \mathscr{L}_{rob}^{sp}$, which encourages the model learning from hard-class pairs.

\subsection{Self-Paced Accuracy Loss} \label{accuracy loss}
According to our empirical observations and theoretical analysis in Section 3, untargeted attacks are prone to generate adversaries from hard-class pairs. We argue that a model with stronger HCP discrimination capability would be more robust against adversarial attacks. To this end, we propose up-weighting HCP loss and down-weighting ECP loss in model training to facilitate discriminative representation learning. 

As shown in the analysis in Section 3.2, $\cos(\boldsymbol{\theta}_{j})$ evaluates the representation similarity between $\boldsymbol{z_{x}}$ and the prototype vector $\boldsymbol{w_{j}}$ of the $j^{th}$ class. Ideally, for data from the $i^{th}$ category, we target $\cos(\boldsymbol{\theta}_{j})=\delta(i-j)$, where $\delta(x)$ is the Dirichlet identity function. Toward this goal, we monitor the values of $\cos(\boldsymbol{\theta}_{j})$ and take them as metrics to adaptively re-weight training samples in adversarial training. 

Formally, we propose to reshape the NCE loss by the self-paced modulating factors $g^{t}$ and $g^f_j$:
\begin{equation}
    \mathscr{L}_{acc}^{sp} = -\log(\frac{e^{g^t\boldsymbol{w_{i}}^T\boldsymbol{z_{x}}}}{\sum_{j\neq i}^{C}e^{g^f_j \boldsymbol{w_{j}}^T \boldsymbol{z_{x}}}+e^{g^t\boldsymbol{w_{i}}^T\boldsymbol{z_{x}}}}),
    \label{NCE_SP}
\end{equation}
where $\lVert \boldsymbol{w_{i}}\rVert_2=1$ and $\lVert \boldsymbol{z_{x}}\rVert_2=s$ ~\cite{wang2018cosface}. For a sample with true label $i$, the true-class modulating gain $g^{t}$ and false-class weights $g^f_j$ are defined as
\begin{equation}
\label{wsp}
    \begin{cases}
      g^t = 1-\cos(\boldsymbol{\theta}_{i})+\beta\\
      g^{f}_j = \cos(\boldsymbol{\theta}_{j}) + \beta\\     
    \end{cases}.
\end{equation}
$\beta$ is a smoothing hyper-parameter to avoid $g^t=0$ and $g^f_j=0$. This study adopts the NCE loss, rather than the CE loss, in $\mathscr{L}_{acc}^{sp}$ for the following reasons. NCE is a hypersphere embedding. Compared to the CE loss, the directional embedding encourages a model to learn more discriminative features~\cite{wang2018cosface, liu2017sphereface, schroff2015facenet}. Recent study in \cite{pang2020boosting} further shows that deploying NCE in adversarial training boosts model robustness against various attacks. It is noteworthy that our ablation study shows that the proposed self-paced modulating mechanism does not only boost model robustness with the NCE loss but also improves model performance with the CE loss. 

Intuitively, the introduced self-paced modulating factors amplify the loss contribution from hard-class pairs, and meanwhile down-weight easy-class pair loss. Specifically, according to (\ref{wsp}), data from the $i^{th}$ category are associated with large $g^t$ and $g^f_j$ when its representation $z_x$ is far away from its true-class prototype vector $\boldsymbol{w_{i}}$ while close to a false-class prototype $\boldsymbol{w_{i}}$. In this scenario, $z_x$ and a false-class prototype vector $\boldsymbol{w_{j}}$ constitutes a hard-class pair and both $g^t$ and $g^f_j$ amplify the loss in (\ref{NCE_SP}), encouraging the model to learn a better representation. On the other hand, when $z_x$ and a false-class prototype vector $\boldsymbol{w_{j}}$ constitutes an easy-class pair with small $\cos(\boldsymbol{\theta}_{j})$, $g^{f}_j$ is small and thus reduces the ECP contributions to model optimization.

\subsection{Self-Paced Robustness Loss}
The robustness loss term in AT encourages a model to generate the same label to both clean data $x$ and their adversarial samples $x'$. Intuitively, given a robust representation model, $x$ and $x'$ should share the same hard-class pairs and easy-class pairs. From our analysis in Section 3.2, such an HCP-ECP consistency constraint on $x$ and $x'$ can be formulated as:
\begin{equation}
\cos(\boldsymbol{\theta}_{j}) \approx \cos(\boldsymbol{\theta'}_{j}), \forall{j}.
\label{robustHCP}
\end{equation}
$\boldsymbol{\theta'}_{j}$ is the angle between $z_{x'}$ and a prototype vector $\boldsymbol{w_{j}}$ in the softmax layer of a model.

In prior arts, KL divergence is a widely used as a surrogate robust loss in AT \cite{wang2019improving,zhang2019theoretically}. It quantifies the difference between predicted logits on clean data and its adversarial version:
\begin{equation} \label{eqn:KL}
    KL(f(x)\rVert f(x')) = \sum_{i=1}^Cf_{i}(x)log\frac{f_{i}(x)}{f_{i}(x')}.
\end{equation}
Though the $KL$ divergence measures the logit similarity from the point of view of statistics, it doesn't impose the aforementioned HCP-ECP consistency constant in (\ref{robustHCP}) on model optimization. 

In this study, we propose a new regularization factor, $L_{inc}(x,x')$, to penalize HCP-ECP inconsistency in model robustness training. With simple math, (\ref{robustHCP}) can be converted into a more intuitive expression: $f_j(x)\approx f_j(x')$ for all $j$. To accommodate the two inconsistency conditions, $f_j(x)\gg f_j(x')$ and $f_j(x)\ll f_j(x')$), within one formula, we propose the use of $[log \frac{f_j(x)}{f_j(x')}]^2$ to quantify the HCP-ECP inconsistency between $x$ and $x$ with respect to a specific class $j$. Another benefit of the square operation is its amplification effect on large values, which encourages the model to satisfy the HCP-ECP consistency constraint. Instead of accumulating all inconsistency penalties direction, we follow the statistic perspective of computing KL divergence and the new regularization factor is formulated as
\begin{equation} \label{eqn:Linc}
    L_{inc}^{sp}(x,x') = \sum_j^C [f_j(x)log \frac{f_j(x)}{f_j(x')}]^2.
\end{equation}
Therefore, our new robustness loss is
\begin{equation} \label{eqn:spRobust}
    \mathscr{L}_{rob}^{sp} = \alpha KL(f(x)\rVert f(x')) + L_{inc}^{sp}(x,x'),
\end{equation}
where $\alpha$ is a hyper-parameter to balance the two robustness terms.


\section{Experiments}
In this section, we first conduct a comprehensive empirical study on the proposed SPAT, providing an in-depth analysis of the method. Then we evaluate SPAT on two popular benchmark datasets, MNIST and CIFAR10, in both white-box and black-box settings. A comparison study with state-of-the-art AT methods is presented. 

\subsection{Breaking Down SPAT} \label{setting}
To gain a comprehensive understanding of SPAT, three sets of ablation experiments are conducted: (1) Sensitivity to hyper-parameters, (2)Removing the SP factors in the SPAT loss, and (3) Replacing NCE with CE in SPAT. 

\subsubsection{Experimental Setup}. We use ResNet-18~\cite{he2016deep} as our classifier for the CIFAR-10 dataset. Our experimental settings follow prior arts in~\cite{zhang2019theoretically, wang2019improving}. All models in this ablation study are trained 100 epochs with SGD and the batch size is 128. The initial learning rate is set as 0.1 and decays by 10 times at $75^{th}$ and $90^{th}$ epoch. At the training stage, we use 10-step PGD to generate adversarial samples, with $\epsilon=8/255$, step size = $\epsilon/4$, and $\lambda=6$. For evaluation, we apply 20-step PGD to generate attack data, with $\epsilon=8/255$,  step size = $\epsilon/10$. The default hyper-parameter in all experiments are $s = 5$ and $\alpha=\beta=0.2$, unless otherwise specified.


\begin{table*}[t]
\centering
\subfloat[Varying $s$ in $\mathscr{L}_{acc}^{sp}$]{
\centering
\begin{tabular}{c|c|c}
    \hline
    $s$ & Clean & PGD-20\\
    \hline
    \hline
    1  & \textbf{87.57} & 49.52\\
  \hline
    3  & 86.16 & 55.77\\
  \hline
   5  & 84.26 & 59.56\\
  \hline
   8  & 82.54 & 60.24\\
  \hline
   10 & 81.24 & \textbf{61.02}\\
  \hline
\end{tabular}
\label{a}
}
\hspace{0.6 cm}
\subfloat[Varying $\alpha$ in $\mathscr{L}_{rob}^{sp}$]{
\centering
\begin{tabular}{c|c|c}
\hline
     $\alpha$ & Clean & PGD-20\\
\hline
\hline
  0.0 & \textbf{84.66} & 58.32\\
  \hline
   0.2  & 84.26 & 59.56\\
   \hline
   0.4  &83.60 & 60.11\\
   \hline
   0.6 &83.01 & \textbf{60.57}\\
   \hline
    \hline
\end{tabular}
\label{b}
}
\hspace{0.6 cm}
\subfloat[Varying $\beta$ in $\mathscr{L}_{acc}^{sp}$]{
\centering
\begin{tabular}{c|c|c}
\hline
     $\beta$ & Clean & PGD-20\\
\hline
\hline
  0.0 & \textbf{85.03} & 57.88\\
  \hline
   0.2  & 84.26 & 59.56\\
   \hline
   0.4  &83.81 & \textbf{59.64}\\
   \hline
   0.6 &82.66 & 57.62\\
   \hline
\end{tabular}
\label{c}}
\caption{Hyper-parameter sensitivity in SPAT. If unspecified, the default values are: $s=5, \alpha=\beta=0.2$.}
\end{table*}

\subsubsection{Sensitivity of Hyper-parameters} 
SPAT has three newly introduced hyper-parameters, $s$ and $\alpha$ in $\mathscr{L}_{acc}^{sp}$ and $\beta$ in $\mathscr{L}_{rob}^{sp}$. Table 1 presents the sensitivity of these hyper-parameters on CIFAR-10 dataset and shows their impacts on model accuracy and robustness. The best performance metrics are highlighted in bold. Similar to NCE\cite{wang2018cosface,pang2020boosting}, the scale factor $s$ in SPAT regulates the length of embeddings. From Table~\ref{a}, a larger $s$ leads to higher robustness but lower accuracy. This is because a larger $s$ indicates a larger spherical embedding space and thus samples from different classes can be distributed more discretely. However, the relatively-sparse sample distribution in the large embedding space increases the difficulty of classification. 
$\alpha$ and $\beta$ are parameters up-weighting hard-class pair loss in SPAT. As shown in Table~\ref{b} and \ref{c}, appropriately choosing $\alpha$ and $\beta$ can boost model robustness with little accuracy degradation.

\subsubsection{Analysis of SP:} Table~\ref{break down} records the performance when removing the proposed self-paced factors in the SPAT loss function. Note, when removing SP weights in the accuracy loss, we let $g^t=g_j^f=0$ and the proposed self-paced NCE loss becomes the original NCE loss. 
As indicated in Table~\ref{break down}, removing the SP mechanism from either robustness loss or accuracy loss leads to substantial performance degradation. In particular, the introduced self-paced robustness term encourages the model to follow the HCP/ECP consistency constraint, which contributes to a larger margin of robustness improvement. 

\subsubsection{Analysis of NCE in SPAT:} This study introduces the self-paced modulation factors upon the NCE loss. Table~\ref{Normalization} compares model performance when we replace NCE with either the CE loss or a self-paced CE loss (by relaxing normalization $v_j=1$). The normalization regularization in NCE boosts both model robustness and standard accuracy. In addition, incorporating the self-paced factors into the CE loss also improves model performance. This observation validates our innovation of up-weighting hard-class pair loss in model optimization. 

\begin{table}[t]
    \centering
    \begin{tabular}{c|c|c}
    \hline
        loss functions & Clean & PGD-20\\
        \hline
        \hline
        $L_{nce}^{sp} + \lambda L_{rob}^{sp}$ & \textbf{84.26} & \textbf{59.56} \\
        \hline
        $L_{nce} + \lambda L_{rob}^{sp}$& 82.49 & 58.74\\
         \hline
        $L_{nce}^{sp} + \lambda L_{rob}$ & 84.01 & 56.14\\
         \hline
        $L_{nce} + \lambda L_{rob}$& 83.33 & 54.58\\
         \hline
    \end{tabular}
    \caption{Removing SP factors from SPAT.}
    \label{break down}
\end{table}

\begin{table}[t]
    \centering
    \begin{tabular}{c|c|c}
    \hline
        loss functions & Clean & PGD-20\\
        \hline
        \hline
         $L_{nce}^{sp} + \lambda L_{rob}^{sp}$ & \textbf{84.26} & \textbf{59.56} \\
         \hline
         $L_{ce}^{sp} + \lambda L_{rob}^{sp}$ & 82.86 & 53.55 \\
         \hline
        $L_{ce} + \lambda L_{rob}$ & 82.12 & 51.82 \\
        \hline
    \end{tabular}
    \caption{Replacing NCE with CE in SPAT.}
    \label{Normalization}
\end{table}

\begin{table}
\centering
\begin{tabular}{ c||c||c|c|c} 
 \hline
 defense & Clean & FGSM & PGD-20 & C\&W\\
 \hline
 \hline
    Madry's& 99.15 & 97.22 & 95.51 & 95.66\\
  \hline
    ALP& 98.79 & 97.31 & 95.85 & 95.50\\
  \hline
   TRADES& 99.10 & 97.42 & 96.22 & 96.01\\
  \hline
   MART& 98.89 & 97.70 & 96.24 & 96.33\\
  \hline
   SPAT& \textbf{99.21}&\textbf{98.12}& \textbf{96.64}& \textbf{96.57}\\
  \hline
\end{tabular}
\caption{White box robustness accuracy(\%) on MNIST}
\label{white box mnist}
\centering
\begin{tabular}{ c||c||c|c|c} 
 \hline
 defense & Clean & FGSM & PGD-20 & C\&W\\
 \hline
 \hline
    Madry's& 99.15 & 97.06 & 96.00 & 96.88\\
  \hline
    ALP& 98.79 & 97.23 & 96.13 & 97.32\\
  \hline
   TRADES& 99.10 & 97.27 & 96.88 & 97.03\\
  \hline
   MART& 98.89 & 97.68 & 96.73 & 97.20\\
  \hline
   SPAT& \textbf{99.21}&\textbf{97.80}& \textbf{97.27}& \textbf{97.40}\\
  \hline
\end{tabular}
\caption{Black box robustness accuracy(\%) on MNIST.}
\label{black box mnist}
\end{table}

\begin{table*}[h]
\parbox{1\linewidth}{
\centering
\begin{tabular}{ c||c||c|c|c|c|c|c|c} 
 \hline
 defense & Clean & FGSM & PGD-20 & PGD-100 & MIM-20 & FAB & C\&W & AA\\
 \hline
 \hline
    Madry's&\textbf{84.35}&54.23&46.70&45.73&47.03&47.67&48.62&46.90\\
  \hline
   TRADES& 82.12& 56.49& 51.82& 50.21& 51.25& 48.21&49.96&47.32 \\
  \hline
   MART& 83.08& 60.19 & 54.87 & 52.97& 53.91 & 48.62 & \textbf{51.23} & 47.87 \\
  \hline
   GAIRAT& 83.14& 60.03 & 54.85 & 52.68& 53.44 & 37.11 & 40.73 & 35.90 \\
   \hline
   MAIL-AT& 83.80& 61.33 & 55.06 & 53.26& 54.57 & 45.55 & 48.67 & 44.32 \\
   \hline
   SEAT & 83.20& 61.54 & 55.86 & 55.53& 57.01 & 45.70 & 49.03 & 47.43 \\
   \hline
   SPAT& 84.08&\textbf{61.71}& \textbf{58.33}& \textbf{58.11}& \textbf{58.93}& \textbf{48.54} &50.60&\textbf{48.09}\\
  \hline
\end{tabular}
\caption{White box robustness accuracy(\%) on CIFAR-10 with ResNet-18.}
\label{white box}
}

\parbox{1\linewidth}{
\centering
\begin{tabular}{ c||c||c|c|c|c|c|c|c} 
 \hline
 defense & Clean & FGSM & PGD-20 & PGD-100 & MIM-20 & FAB & C\&W & AA\\
 \hline
 \hline
    Madry's&\textbf{84.35}&79.84&80.35&80.91&80.12&81.93&79.98&82.02\\
  \hline
   TRADES& 82.12& 79.98& 80.69& 80.80& 80.24& 81.71&80.55 &81.91 \\
  \hline
   MART& 83.08& 81.50 & 82.31 & 82.89& 82.04 & 83.02 & 82.97 & 83.06 \\
   \hline
   GAIRAT& 83.14& 79.92 & 80.40 & 80.61& 80.22 & 82.49 & 82.43 & 82.69 \\
   \hline
   MAIL-AT& 83.80& 81.22 & 82.16 & 82.37& 81.96 & 83.10 & 82.38 & 83.36 \\
   \hline
   SEAT& 83.20& 80.44 & 81.60 & 82.15& 82.33 & 83.05 & 81.90 & 83.10 \\
  \hline
   SPAT& 84.08&\textbf{82.39}& \textbf{83.20}& \textbf{83.41}& \textbf{82.91}& \textbf{83.98} &\textbf{84.05}&\textbf{84.07}\\
  \hline
\end{tabular}
\caption{Black box robustness accuracy(\%) on CIFAR-10 with ResNet-18.}
\label{black box}
}
\end{table*}

\subsection{Robustness Evaluation under Different Attacks }

In this section, we evaluate the robustness of SPAT on two benchmarks, MNIST and CIFAR10, under various attacks. 

\subsubsection{Experimental settings:} For MNIST, we use a simple 4-layer-CNN followed by three fully connected layers as the classifier. We apply 40-step PGD to generate adversaries in training, with $\epsilon=0.3$ and step size of 0.01. We train the models for 80 epochs with the learning rate of 0.01. Since MNIST is a simple dataset, three classical attacks, FGSM~\cite{goodfellow2014explaining}, PGD-20~\cite{madry2017towards}, and C\&W with $l_{\infty}$ ~\cite{carlini2017adversarial}, are deployed in our white-box and black-box settings.

On CIFAR-10, adversarial samples used in ATs are generated by 10-step PGD, with $\epsilon = 8/255$ and step size of $\epsilon/4$. The rest training setup is the same as in section~\ref{setting}. Since CIFAR-10 is a more complex dataset, we further include four stronger attacks in this experiment, which are PGD-100, MIM~\cite{dong2018boosting}, FAB~\cite{croce2020minimally}, and AutoAttack (AA)~\cite{croce2020reliable}. All attacks are bounded by the $l_\infty$ box with the same maximum perturbation $\epsilon=8/255$. 

\subsubsection{Baselines:} SOTA defense methods including Madry's~\cite{madry2017towards}, TRADES~\cite{zhang2019theoretically}, MART~\cite{wang2019improving}, GAIRAT~\cite{zhang2020geometry}, MAIL-AT~\cite{liu2021probabilistic} and SEAT~\cite{wang2022self} are evaluated in this comparison study. We follow the default hyperparameter settings presented in the original papers. For instance, $\lambda=6$ in TRADES and 5 in MART. For ALP, we set the weight for logit paring as 0.5. 

\subsubsection{White-Box Robustness:} 

Table.~\ref{white box mnist} and Table~\ref{white box} report the white-box robustness performance on MNIST and CIFAR-10, respectively. We omit the standard deviations of 4 runs as they are typically small ($<$ 0.50$\%$), which hardly affects the results. SPAT achieves the highest robustness in all 4 attacks on MNIST and 6 out of 7 on CIFAR-10. The only exception is the $l_\infty$ C\&W attack which directly optimizes the difference between correct and incorrect logits~\cite{madry2017towards}. Notice that the optimization function of the C\&W attack ($l_\infty$ version) is the same as the objective function (boosted cross entropy) for MART which makes the rest defense strategies in an unfair position. Even so, SPAT is only 0.67\% less robust than MART under the C\&W attack. We shown in Appendix that the proposed SPAT also works well with larger models such as WideResNet-34. 

\subsubsection{Black-Box Robustness:}
In the black-box attack setting, since adversaries do not access the model architecture and parameters, adversarial samples are crafted on a naturally trained model and transferred to the evaluated models. Here we use a naturally trained LENET-5~\cite{lecun1998gradient} and ResNet101 for adversarial sample generation, whose natural accuracy is 98.94\% and 95.53\% on MNIST and CIFAR-10 respectively. 

Table.~\ref{black box mnist} and Table~\ref{black box} report the white-box robustness performance on MNIST and CIFAR-10, respectively.
Since the features for MNSIT is simple and linear, we notice for certain cases the black box attacks are even stronger than the white box attacks. For example, white box FGSM attacks are weaker than their black box counterpart on all defenses. On the CIFAR10 dataset, while all models reach much higher robustness accuracy compared to white box attacks, SPAT again achieves the top performance. It is worth noting that the weakest attack (FGSM) has the highest black box transferability, while the strongest attack method, AutoAttack, has almost no effect on the SPAT trained model (from 84.08\% to 84.07\%).

In addition, our experimental results on CIFAR-10C in Appendix suggest that the model trained by SPAT is also robust to natural image corruptions.

\section{Conclusion and Future work}
In this paper, we studied an intriguing property of untargeted adversarial attacks and concluded that the direction of a first-order gradient-based attack is largely influenced by its hard-class pairs. With this insight, we introduced a self-paced adversarial training strategy and proposed up-weighting hard-class pair loss and down-weighting easy-class pair loss in model optimization. Such an online re-weighting strategy on hard/easy-class pairs encouraged the model to learn more useful knowledge and disregard redundant, easy information.
Extensive experiment results show that SPAT can significantly improve the robustness of the model compared to state-of-the-art AT strategies. 

In the future, on one hand, we plan to apply the hard/easy-class pair re-weighting principles to recently proposed AT algorithms, and explore the potential improvement by differentiating hard/easy-class pairs in AT. On the other hand, we plan to investigate "true" untargeted adversarial attacks so that the adversarial perturbations are less predictable. 
\newpage
{\small
\bibliography{egbib}
}

\clearpage
\appendix
\setcounter{secnumdepth}{0}

\onecolumn
\section{Supplementary material}

\vspace{0.3cm}
The supplementary material of our paper, entitled "Improving Adversarial Robustness with Self-Paced Hard-Class Pair Reweighting", includes this technical appendix and our SPAT source code.

\vspace{0.3cm}
\subsection{A. Proof of Lemma 1}
\subsubsection{Lemma 1:} \textit{For an oracle model that predicts the labels perfectly on clean data, the gradient of the CE loss with respect to sample $x$ from the $i^{th}$ category is:}
\begin{equation}
\nabla_{\boldsymbol{x}}\mathscr{L}(f(x),y)=[\sum_{j\neq i}^{C}\sigma(\boldsymbol{w_{j}}^{T}\boldsymbol{z_x})\boldsymbol{w_{j}}]\nabla_{\boldsymbol{x}}\boldsymbol{z_x}, \nonumber
\end{equation}
where $\sigma(\boldsymbol{w_{i}}^{T}\boldsymbol{z_x}) = \frac{e^{\boldsymbol{w_{i}}^{T}\boldsymbol{z_x}}}{\sum_{j=1}^{C}e^{\boldsymbol{w_{j}}^{T}\boldsymbol{z_x}}}$.

\subsubsection{Proof:} The CE loss can be formulated as
\begin{equation}
\mathscr{L}(f(x),y)=-log\frac{e^{\boldsymbol{w_{i}}^{T}\boldsymbol{z_x}}}{\sum_{j=1}^{C}e^{\boldsymbol{w_{j}}^{T}\boldsymbol{z_x}}}. \nonumber
\end{equation}
Hence,
\begin{equation}
\begin{aligned}
\nabla_{\boldsymbol{x}}\mathscr{L}(f(x),y)
&= -\nabla_{\boldsymbol{x}}log\frac{e^{\boldsymbol{w_{i}}^{T}\boldsymbol{z_x}}}{\sum_{j=1}^{C}e^{\boldsymbol{w_{j}}^{T}\boldsymbol{z_x}}} \\
&= -\nabla_{\boldsymbol{x}}[log e^{\boldsymbol{w_{i}}^{T}\boldsymbol{z_x}}-log \sum_{j=1}^{C}e^{\boldsymbol{w_{j}}^{T}\boldsymbol{z_x}}] \\
&= -\nabla_{\boldsymbol{x}}[\boldsymbol{w_{i}}^{T}\boldsymbol{z_x}]+\nabla_{\boldsymbol{x}}[log \sum_{j=1}^{C}e^{\boldsymbol{w_{j}}^{T}\boldsymbol{z_x}}]\\
&= -\nabla_{\boldsymbol{x}}[\boldsymbol{w_{i}}^{T}\boldsymbol{z_x}]+\frac{1}{\sum_{k=1}^{C}e^{\boldsymbol{w_{k}}^{T}\boldsymbol{z_x}}}\nabla_{\boldsymbol{x}}[\sum_{j=1}^{C}e^{\boldsymbol{w_{j}}^{T}\boldsymbol{z_x}}]\\
&= -\nabla_{\boldsymbol{x}}[\boldsymbol{w_{i}}^{T}\boldsymbol{z_x}]+\frac{1}{\sum_{k=1}^{C}e^{\boldsymbol{w_{k}}^{T}\boldsymbol{z_x}}} \cdot\sum_{j=1}^{C}e^{\boldsymbol{w_{j}}^{T}\boldsymbol{z_x}}\nabla_{\boldsymbol{x}}[\boldsymbol{w_{j}}^{T}\boldsymbol{z_x}]\\
&= -\nabla_{\boldsymbol{x}}[\boldsymbol{w_{i}}^{T}\boldsymbol{z_x}]+\sum_{j=1}^{C}\frac{e^{\boldsymbol{w_{j}}^{T}\boldsymbol{z_x}}}{\sum_{k=1}^{C}e^{\boldsymbol{w_{k}}^{T}\boldsymbol{z_x}}}\nabla_{\boldsymbol{x}}[\boldsymbol{w_{j}}^{T}\boldsymbol{z_x}]\\
&= -\nabla_{\boldsymbol{x}}[\boldsymbol{w_{i}}^{T}\boldsymbol{z_x}]+\sum_{j=1}^{C}\sigma(\boldsymbol{w_{j}}^{T}\boldsymbol{z_x})\nabla_{\boldsymbol{x}}[\boldsymbol{w_{j}}^{T}\boldsymbol{z_x}]\\
&= [\sigma(\boldsymbol{w_{i}}^{T}\boldsymbol{z_x})-1]\boldsymbol{w_{i}}^{T}\nabla_{\boldsymbol{x}}\boldsymbol{z_x}+[\sum_{j\neq i}^{C}\sigma(\boldsymbol{w_{j}}^{T}\boldsymbol{z_x})\boldsymbol{w_{j}}]\nabla_{\boldsymbol{x}}\boldsymbol{z_x}\\ \nonumber
\end{aligned}
\end{equation}
For an oracle model that predicts the labels perfectly on clean data, $\sigma(\boldsymbol{w_{i}}^{T}\boldsymbol{z_x})=1$ for a data from the $i^{th}$ class. Hence, the first term in the proof vanishes. That is,
\begin{equation}
\nabla_{\boldsymbol{x}}\mathscr{L}(f(x),y)=[\sum_{j\neq i}^{C}\sigma(\boldsymbol{w_{j}}^{T}\boldsymbol{z_x})\boldsymbol{w_{j}}]\nabla_{\boldsymbol{x}}\boldsymbol{z_x}, \nonumber
\end{equation}

\vspace{0.2cm}
\subsection{B. Statistics of weight norms of the softmax layer in CE-trained models}
We calculate the Euclidean norms of weights from the softmax layer in CE-trained models on CIFAR-10 and report the values in Table \ref{WeightNorm}.

\begin{table}[h]
\centering
\begin{tabular}{ c||c||c|c} 
 \hline
 Models  & ResNet-18 & ResNet-34 & ResNet-50\\
 \hline
 \hline
    $v_0$& 1.113 & 1.036 & 1.072\\
  \hline
   $v_1$& 1.134 & 1.063 & 1.106 \\
  \hline
   $v_2$& 1.103 & 1.023 & 1.055 \\
  \hline
  $v_3$& 1.082 & 1.005 & 1.032 \\
  \hline
   $v_4$& 1.113 & 1.037 & 1.068 \\
   \hline
   $v_5$& 1.096 & 1.023 & 1.057 \\
   \hline
   $v_6$& 1.122 & 1.048 & 1.083 \\
   \hline
   $v_7$& 1.121 & 1.046 & 1.085 \\
   \hline
   $v_8$& 1.127 & 1.051 & 1.096 \\
  \hline
   $v_9$& 1.121& 1.046 & 1.088\\
  \hline
\end{tabular}
\caption{Weight norms of the softmax layer in CE-trained models on CIFAR-10}
\label{WeightNorm}
\end{table}

\vspace{0.3cm}
\subsection{C. Pseudocode of SPAT}
The pseudocode of the proposed SPAT algorithm is presented in Algorithm \ref{alg:cap}.
\begin{algorithm*}[]
\caption{Self-Paced Adversaral Training}\label{alg:cap}
\begin{algorithmic}[1]
\State \textbf{Input:} Number of training data $N$, batch size $m$, number of iterations for inner optimization $K$, maximum perturbation $\epsilon$, step sizes $\eta_1$ and $\eta_2$, classifier parameterized by $\theta$
\State \textbf{Output:} Robust classifier $f_\theta$
\For{$s$ = 1... $N/m$}
\State read mini-batch \{$\boldsymbol{x_1}, ..., \boldsymbol{x_m}$\} from training data
    \For{$i$ = 1... $m$}
    \State $\boldsymbol{x'_i} = \boldsymbol{x_i} + 0.001 * \boldsymbol{\mathcal{N}(0, I)}$  \Comment{$\mathcal{N}(0, I)$ is the standard normal distribution}
        \For{$j$ = 1... $K$}
            \State $\boldsymbol{x'_i} = \Pi_{\mathscr{B}(\boldsymbol{x_i}, \epsilon)}(x'_i +  \eta sign(\nabla_{\boldsymbol{x}}\mathscr{L}_{rob}^{sp}(\boldsymbol{x_i}, \boldsymbol{x'_i}; \boldsymbol{\theta}))$ \Comment{$\Pi$ is the projection operator}
        \EndFor
    \EndFor
    \State $\boldsymbol{\theta} = \boldsymbol{\theta} - \eta_2\sum_{i}^m \nabla_{\boldsymbol{\boldsymbol{\theta}}}\mathscr{L}^{SPAT}(\boldsymbol{x_i}, \boldsymbol{y_i}, \boldsymbol{x'_i}; \boldsymbol{\theta})$
\EndFor
\end{algorithmic}
\end{algorithm*}

\vspace{0.2cm}

\subsection{D. Robustness Evaluation with Larger Models}
To discover the full potential of SPAT, we conduct more experiments on models with larger capacity (WideResNet-34-10~\cite{zagoruyko2016wide}) and compare them with state-of-the-art defenses. The scale factor $s$ is set to 9 to adapt larger model capacity. The rest of the training settings are the same as those used in section~\ref{setting}. The results are shown in Table.~\ref{WRN}, where SPAT still achieves top robustness performance.

\begin{table}[h]
\centering
\begin{tabular}{ c||c||c|c|c} 
 \hline
 defense & Clean & PGD-100 & C\&W & AA\\
 \hline
 \hline
    Madry's*& \textbf{87.80} & 49.43 & 53.38 & 48.46\\
  \hline
   TRADES*& 86.36 & 54.88 & 56.18 & 53.40\\
  \hline
   MART*& 84.76 & 55.61 & 54.72 & 51.40\\
  \hline
  GAIRAT*& 86.30 & 58.74 & 45.57 & 40.30\\
  \hline
   MAIL-AT*& 84.83 & 58.86 & 51.26 & 47.10\\
  \hline
   SPAT& 85.13&\textbf{60.44}& \textbf{55.60}& \textbf{54.92}\\
  \hline
\end{tabular}
\caption{White box robustness accuracy(\%) on CIFAR-10 with WideResNet-34-10. * Results are directly from \cite{liu2021probabilistic}.}

\label{WRN}
\end{table}

\subsection{E. Case Study: Naturally Corrupted Perturbation}
Adversarial attack is the most extreme scenario for evaluating the robustness of models. Unlike adversarial attacks, naturally-corrupted data, such as blurring, compression, defocusing, \textit{etc}, do not require model information to generate noises and can be seen as a type of generic black-box attack. 
In this experiment, we explore the potential of SPAT on such naturally-corrupted data. We apply the SPAT-trained ResNet-18 in Section 5.2 on the corrupted CIFAR-10 dataset (\textit{CIFAR-10-C}~\cite{hendrycks2019robustness}). In CIFAR-10-C, the clean CIFAR-10 data are processed to mimic various image distortions under harsh conditions. Table~\ref{corr benchmark} presents classification accuracy on the CIFAR-10-C dataset. Results show that the SPAT-trained model exhibits stronger robustness to different types of corruption. 

\begin{table}[h]
\centering
\begin{tabular}{ c||c|c|c|c} 
 \hline
 defense & PGD & TRADES & MART & SPAT\\
 \hline
 \hline
    Blur& 73.23 & 72.43 & 73.35 &\textbf{ 74.71}\\
  \hline
   Contrast& 78.68 & 76.05 & 76.59 & \textbf{78.39} \\
  \hline
   Fog& 46.38 & 45.74 & 45.19 & \textbf{49.34}\\
  \hline
   Frost&  70.59 & 64.65 & 70.39 & \textbf{71.36}\\
  \hline
  Snow & 73.95 & 70.08 & 74.92 & \textbf{75.02}\\
  \hline
  jpeg& 81.09 & 79.09 & 80.42 & \textbf{81.73}\\
  \hline
  Saturate& 80.23 & 78.68 & 80.51 & \textbf{81.92}\\
  \hline
  Defocus& 77.66 & 76.05 & 76.59 & \textbf{78.39}\\
  \hline
\end{tabular}
\caption{Accuracy (\%) of different corruption types in CIFAR10-C.}
\label{corr benchmark}
\end{table}

\subsection{F. Ablation Experiments of SPAT}
In this section, we demonstrate the effect of each component in SPAT. By removing both SP and NCE, SPAT decays to the Trades algorithm~\cite{zhang2019theoretically}.
\begin{figure}[h]
  \centering
    \subfloat[
    testing standard accuracy] {\includegraphics[width=0.5\textwidth]{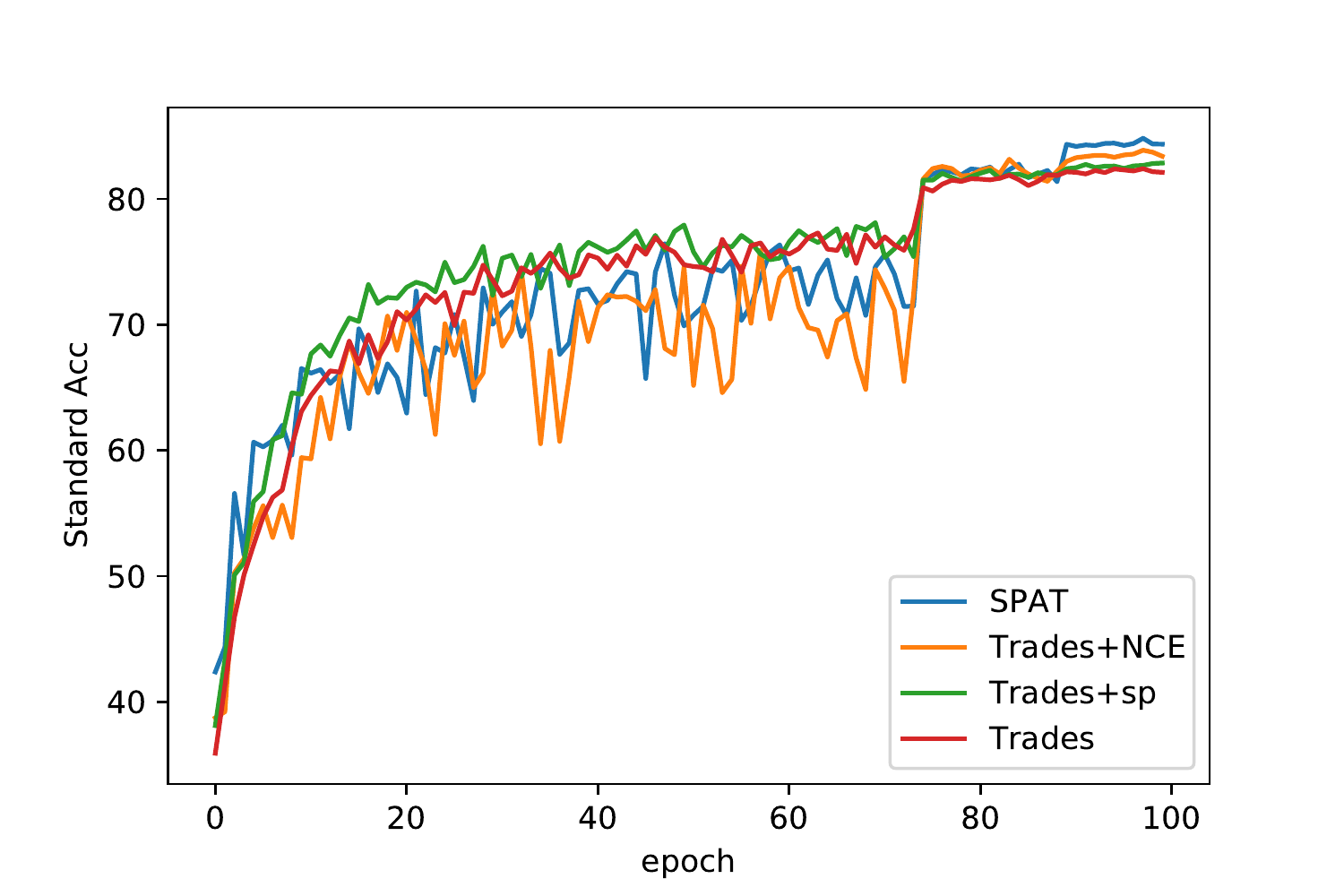}}
  \subfloat[testing robustness accuracy]{\includegraphics[width=0.5\textwidth]{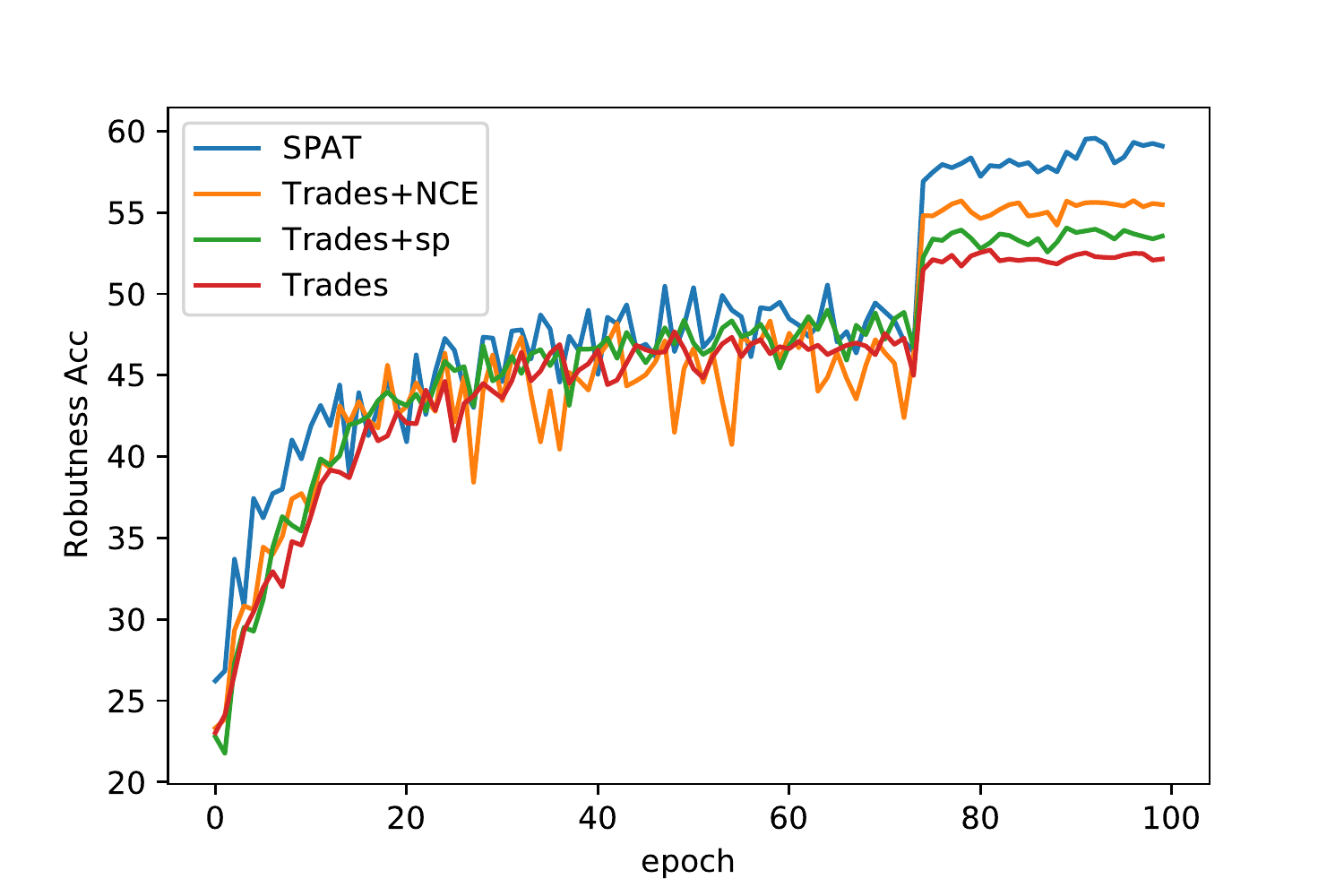}}
  \caption{Ablation experiments corresponding to Table~\ref{break down} and \ref{Normalization}, each component of SPAT can boost the performance against Trades~\cite{zhang2019theoretically}. Combining SP and NCE (SPAT) yields the best performance in standard accuracy and adversarial robustness.}
  \label{ablation}
\end{figure}
\end{document}